\documentclass[10pt,twocolumn,letterpaper]{article}

\usepackage[pagenumbers]{wacv} % To force page numbers, e.g. for an arXiv version

\usepackage{graphicx}
\usepackage{amsmath}
\usepackage{amssymb}
\usepackage{array}
\usepackage{stfloats}
\usepackage{booktabs}
\usepackage[table,xcdraw]{xcolor}
\usepackage[utf8]{inputenc}
\usepackage[T1]{fontenc}
\usepackage{multirow}
\usepackage[accsupp]{axessibility}

\usepackage{adjustbox}

\usepackage[pagebackref,breaklinks,colorlinks]{hyperref}

\usepackage[capitalize]{cleveref}
\crefname{section}{Sec.}{Secs.}
\Crefname{section}{Section}{Sections}
\Crefname{table}{Table}{Tables}
\crefname{table}{Tab.}{Tabs.}

\def\wacvPaperID{5} % *** Enter the WACV Paper ID here
\def\confName{WACV MaCVi Workshop}
\def\confYear{2025}

\newcommand{\figref}[1]{Figure~\ref{fig:#1}}

\begin{document}

\title{AutoFish: Dataset and Benchmark for Fine-grained Analysis of Fish}

\author{Stefan Hein Bengtson$^{1,2}$, Daniel Lehotský$^{1}$, Vasiliki Ismiroglou$^{1,2}$, Niels Madsen$^{3}$\\ Thomas B. Moeslund$^{1,2}$, and Malte Pedersen$^{1,2}$\\\\
$^1$Visual Analysis and Perception Lab, Aalborg University, Denmark\\
$^2$Pioneer Centre for AI, Copenhagen, Denmark\\
$^3$Section of Biology and Environmental Science, Aalborg University, Denmark\\
}
\maketitle

%%%%%%%%% ABSTRACT
\begin{abstract}
   Automated fish documentation processes are in the near future expected to play an essential role in sustainable fisheries management and for addressing challenges of overfishing. In this paper, we present a novel and publicly available dataset named \textit{AutoFish} designed for fine-grained fish analysis. The dataset comprises 1,500 images of 454 specimens of visually similar fish placed in various constellations on a white conveyor belt and annotated with instance segmentation masks, IDs, and length measurements. The data was collected in a controlled environment using an RGB camera. The annotation procedure involved manual point annotations, initial segmentation masks proposed by the Segment Anything Model (SAM), and subsequent manual correction of the masks. We establish baseline instance segmentation results using two variations of the Mask2Former architecture, with the best performing model reaching an mAP of 89.15\%. Additionally, we present two baseline length estimation methods, the best performing being a custom MobileNetV2-based regression model reaching an MAE of 0.62cm in images with no occlusion and 1.38cm in images with occlusion.
   Link to project page: \url{https://vap.aau.dk/autofish/}.
\end{abstract}

%%%%%%%%% BODY TEXT
\section{Introduction}
Earth's marine ecosystems are facing an unprecedented threat, in large part due to overfishing. Among its most significant consequences are habitat destruction, loss of biodiversity, and ecological imbalances in marine environments, profoundly impacting coastal communities worldwide.
Consequently, there has been a surge in research focused on scalable solutions for monitoring marine environments and turning the tide ~\cite{Food_and_Agriculture_Organization2020,ungoals}.

The traditional approach to fisheries management, relying on catch limits and periodic onshore inspections, has proven insufficient in curbing overfishing and ensuring sustainable fisheries and healthy marine environments~\cite{Hold2015}. 
As a result, there is a pressing need to adopt innovative and technology-driven solutions that can effectively address the challenges posed by overfishing.
Real-time automated monitoring of catch compositions could enhance compliance, reduce manual reporting for fishermen, and provide authorities reliable data for sustainable fisheries management~\cite{Beyan2020}.
A streamlined data collection and decision making could foster a transparent and responsible fishing industry, promoting ecological preservation while supporting the socio-economic well-being of coastal communities~\cite{Wolfgang2019}.

\begin{figure}[t]
    \centering
    \begin{minipage}{.5\linewidth}
    \centering
    \includegraphics[width=\linewidth]{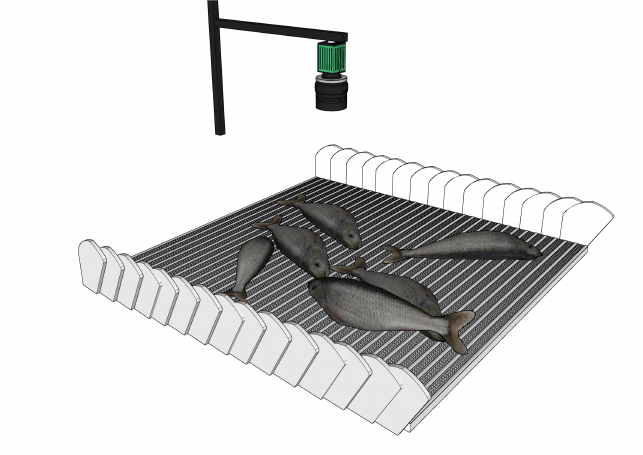}
    \end{minipage} 
    \begin{minipage}{.49\linewidth}
    \centering
    \includegraphics[width=\linewidth]{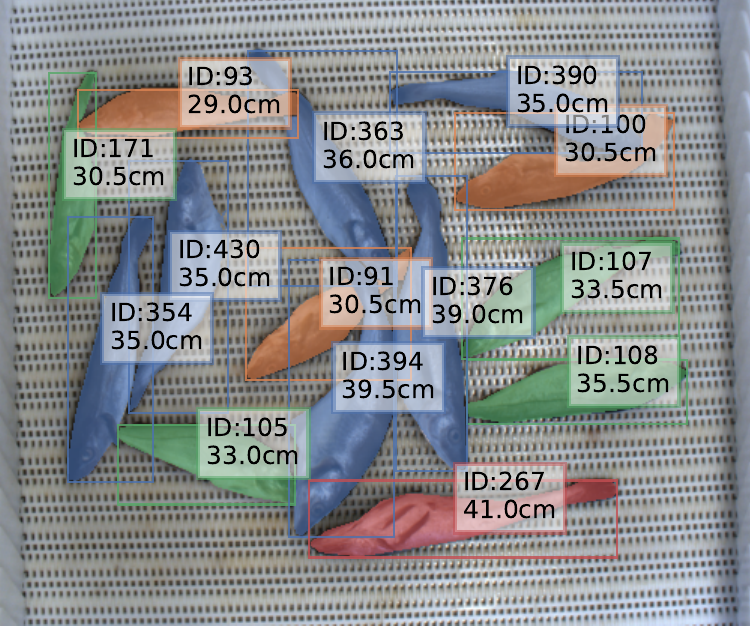}
    \end{minipage}
    \caption{Illustration of the recording setup and an example image from the \textit{AutoFish} dataset with an overlay of groundtruth bounding boxes, instance segmentations, IDs, and lengths.}
    \label{fig:intro}
\end{figure}

In this work, we address the lack of image data from the fishing industry by proposing a novel dataset and baseline results related to automated documentation of catch compositions. 
Our contributions are the following:
\begin{itemize}
    \item \textit{AutoFish}, a publicly available dataset for fine-grained analysis of fish, with instance segmentation masks, IDs and manual length measurements for every specimen. 
    \item A novel group-based data acquisition method for avoiding cross-contamination between data splits.
    \item Baseline results for instance segmentation and length estimation methods on the \textit{AutoFish} dataset.
\end{itemize}

\section{Related work}
\begin{table*}[bp!]
\footnotesize
\centering
\newcolumntype{P}[1]{>{\centering\arraybackslash}p{#1}}
\begin{tabular}{@{}lcP{48pt}P{35pt}P{40pt}P{55pt}ccccccc@{}}
\toprule
\begin{tabular}[c]{@{}c@{}} Dataset\\name\end{tabular}    & Images  & \begin{tabular}[c]{@{}c@{}}Labeling\\type\end{tabular}               & \begin{tabular}[c]{@{}c@{}} Object \\ instances \end{tabular}& \begin{tabular}[c]{@{}c@{}}Instances \\ per image$^*$\end{tabular}       & Location                      & Environment    & IDs    & Length & Weight & \begin{tabular}[c]{@{}l@{}}Both\\sides\end{tabular}\\ \hline
FD  \cite{vanEssen2021}        & 5,231   & Bounding boxes         & 24,008    & 0-30 (4)                & North Sea                     & Conveyor belt  & Yes$^1$& No     & No     & No\\ 
FDWE \cite{Sokolova2023}       & 1,086   & Bounding boxes         & 2,216     & 1-7 (2)                 & North Sea                     & Conveyor belt  & No     & No     & Yes    & No\\ 
Fishnet \cite{kay2021}         & 143,818 & Bounding boxes         & 549,209   & 1-33 (4)$^2$            & Western and Central Pacific   & Vessel deck    & No     & No     & No     & No\\ 
DeepFish \cite{GarciadUrso2022}& 1,320   & Instance segmentation  & 7,339     & 1-29 (7)                & Spain                         & Tray           & No     & Yes    & No     & No\\
\textbf{AutoFish}              & 1,500   & Instance segmentation  & 18,160    & 7-24 (12)               & North Sea                     & Conveyor belt  & Yes    & Yes    & No     & Yes\\ 
\bottomrule
\end{tabular}
\caption{Overview of publicly available and computer vision curated fish catch datasets. $^*$\textit{The number in parentheses is the average number of instances per image.} $^1$\textit{The IDs in the FD dataset were specifically for the purpose of tracking the fish. All reappearances of each fish correspond to the same sequence with no variation in surroundings or orientation.} $^2$\textit{Including humans.}}
\label{tab:datasets}
\end{table*}

Research in automated catch monitoring systems based on computer vision has been conducted for years and, traditionally, species identification and length estimation models have been based on handcrafted features~\cite{Strachan1993,Strachan1994,Zion1999,White2006,Antelo2011}.
However, their design makes them poor at handling unforeseen objects, and it is not trivial to train the models to new species, as they may require an entire new set of handcrafted features or additional parameter tuning.
Recently, there has been a paradigm shift in catch monitoring from handcrafted features to learned features~\cite{Beyan2020}, facilitated by the introduction of effective deep convolutional neural networks (CNNs) for image classification~\cite{AlexNet2012,Girshick2014}.
However, image classification models do not directly take the spatial location of the object into account and is generally not suited for handling scenes with multiple fish.
Nonetheless, image classification has been used for tasks, like identifying the presence of tuna or billfish from electronic monitoring (EM) cameras monitoring the deck of fishing vessels~\cite{Lu2019}.

When handling multiple fish, it is beneficial to know the position (object detection) and species (object classification) of every fish present in the image.
The YOLO~\cite{Redmon2015} architecture has been a popular choice for detecting and classifying fish using bounding box representations.
A YOLOv3 model was used by van Essen et al.~\cite{vanEssen2021} to locate and classify fish on conveyor belts and Sokolova et al.~\cite{Sokolova2023} used a modified YOLOv5 with an additional regression output to detect, classify, and estimate the weight of fish on conveyor belts. However, bounding boxes are not able to accurately capture the shape of non-rigid objects like fish that can bend and deform, or depict occlusion details in cases of overlap. In these cases, it is typically preferred to represent the fish on pixel-level using segmentation masks.
French et al.~\cite{French2015} were among the first to propose methods for monitoring fish on conveyor belts with segmentation masks using CNNs.
They trained and evaluated a Mask R-CNN~\cite{French2019,He2017} model for instance segmentation of round fish, such as haddock, cod, whiting, and hake from EM images. 
Generally, Mask R-CNN is popular in the field and has been used for instance segmentation of fish from EM images~\cite{Tseng2020}, in trawls~\cite{Garcia2019}, in boxes~\cite{Palmer2022}, and in dedicated conveyor belt monitoring systems~\cite{Ovalle2022}.
The latter based their findings on data captured using the iObserver camera system~\cite{Vilas2020} and illustrated the feasibility of using MobileNetV1~\cite{howard2017} for length estimation of fish.

Recent work on length estimation of fish can be divided into two main approaches. The first one being learning-based methods, such as training a small CNN to regress the length of each fish~\cite{Ovalle2022}. The mapping between pixels to real-world lengths, i.e. centimeters, is then indirectly learned by the model.
The other group relies on classical image processing, where pixel-wise lengths are first estimated and then afterwards mapped to real-world lengths.
An example of extracting the pixel-wise length is to extract the central line of the masks for each fish~\cite{Risholm2022} or by identifying key points based on the convex hull of the fish~\cite{Shi2020}.
The final mapping from pixels to centimeters can then be achieved through depth sensors, such as stereo vision~\cite{Shi2020} or time-of-flight~\cite{Risholm2022}.
Another approach to infer the pixels to centimeters mapping is to rely on a reference object with a known size being present in the image, such as an ArUco marker~\cite{Monkman2019}.

Verifying existing findings and driving further development in automated catch monitoring is challenging due to data and annotations typically being kept private.
Additionally, conducting fine-grained analysis is often constrained by the targeted species belonging to distinct families, varying significantly in size, or being positioned in a predictable manner.
Especially the lack of public datasets is a constraint that is becoming increasingly evident with the growing reliance on data-intensive deep learning models.
The few available datasets curated for computer vision tasks are FishNet~\cite{kay2021}, Fish Detection (FD)~\cite{vanEssen2021}, Fish Detection and Weight Estimation (FDWE)~\cite{Sokolova2023}, and DeepFish~\cite{GarciadUrso2022}, as outlined in \cref{tab:datasets} along with our proposed \textit{AutoFish} dataset.
The FishNet dataset~\cite{kay2021} contains images captured from EM cameras on longline tuna vessels in the Pacific.
The images are bounding box annotated and contain a single to a few fish.
Object resolution is generally low due to a wide field of view, which is typical for EM cameras.
The FD~\cite{vanEssen2021} and FDWE~\cite{Sokolova2023} datasets consist of images captured by a dedicated light and camera system mounted above a conveyor belt. 
The images are bounding box annotated and contain information regarding occlusion level, while FDWE also includes the weight of the fish.
The DeepFish dataset~\cite{GarciadUrso2022} contains images of fish on a tray acquired with an iPhone.
The annotations include instance segmentation masks and lengths based on a calibration procedure using the dimensions of the tray on which the fish are placed.

As highlighted in the table, \textit{AutoFish} addresses key gaps within the field, including the lack of instance segmentation and length estimation datasets from conveyor belt environments. 
Additionally, we include IDs and images of both sides of the fish to support more fine-grained analysis.

\section{Dataset}\label{sec:dataset}
We introduce \textit{AutoFish}, a novel and meticulously curated image dataset for fine-grained analysis of fish on a conveyor belt. 
The dataset comprises 1,500 high-quality images featuring 454 unique fish with IDs, manual length measurements, and a total of 18,160 instance segmentation masks. It is to our knowledge the only publicly available dataset of catch that includes multiple documented reappearances of the same fish IDs in different orientations and occlusion levels.
In this section, we provide a detailed account of the process involved in creating the \textit{AutoFish} dataset.

\subsection{Fish composition}
Fish used in the AutoFish dataset were caught and landed by a typically Danish commercial fishing vessel conducting trawl fishery in the North Sea, Skagerrak and Kattegat. The samples mainly consisted of fish species with similar visual characteristics including cod (\textit{Gadus morhua}), haddock (\textit{Melanogrammus aeglefinus}), and whiting (\textit{Merlangius merlangus}). These species are all taxonomic members of the cod family, formally named \textit{Gadidae}. Additionally, hake (\textit{Merluccius merluccius}), and horse mackerel (\textit{Trachurus trachurus}) are well represented in the dataset.

\begin{figure}[t]
    \centering
    \includegraphics[width=\linewidth]{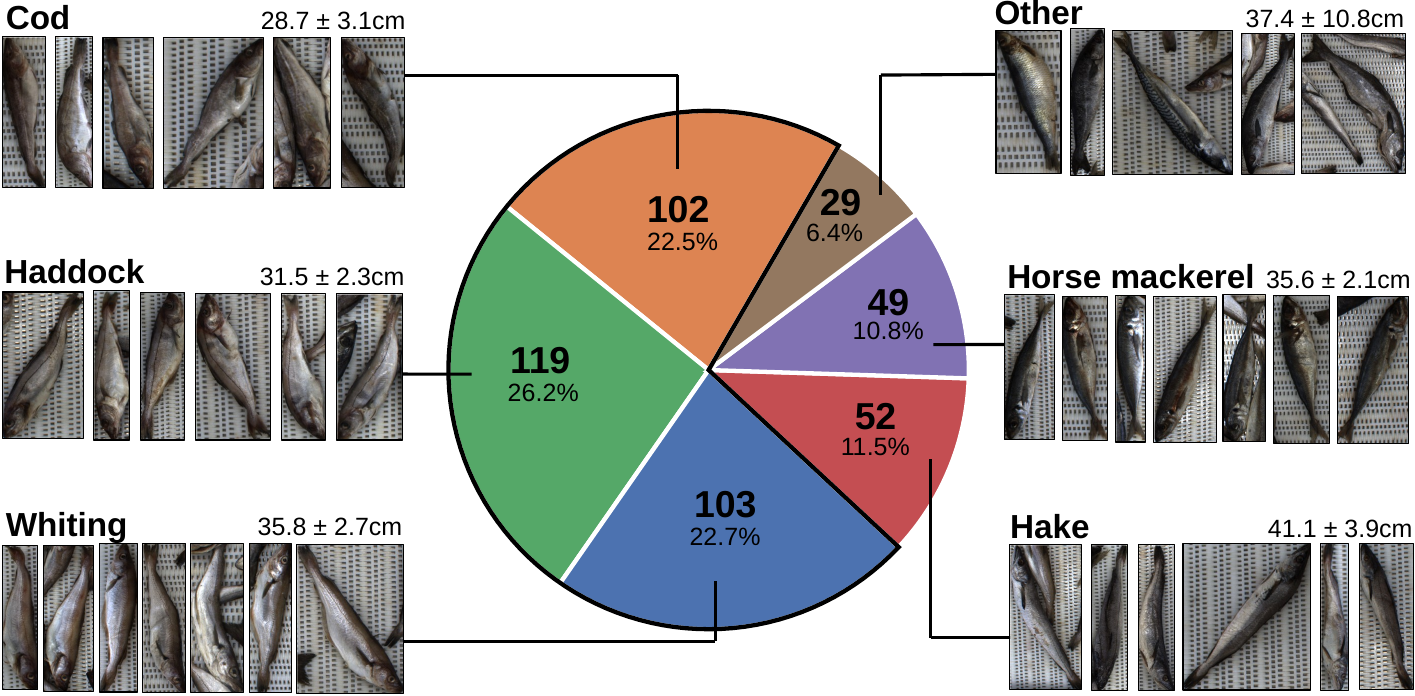}
    \caption{The distribution of species in the AutoFish dataset. The members of the true cod family are highlighted with a black border. The numbers inside the chart indicate the number of specimens. The average length is indicated for each of the species above the image examples.}
    \label{fig:composition}
\end{figure}

All species are of commercial importance and commonly caught in fisheries conducted in the aforementioned areas. 
Therefore, they are essential for developing a dataset that accurately represents the local industry.
An overview of the fish species and the number of individuals included in the \textit{AutoFish} dataset is presented in~\figref{composition}. Note that species represented by only a few individuals are grouped in the \textit{other} category.

\subsection{Camera setup}
The dataset was recorded in a laboratory using a custom setup consisting of a 100x100 cm section of a static white conveyor belt with a camera mounted above, as illustrated in~\figref{intro}. This is similar to the conveyor belt setup that can be commonly found on fishing vessels.
We used a Jai GO-5100C-USB camera, equipped with a KOWA LM12HC lens, and it was placed 1.5 m above the conveyor belt.
The f-stop and focus distance were set to f/11 and 1 m, respectively, to ensure sharp details across the entire image.
The camera was positioned such that the field of view matched the conveyor belt.
The images were recorded in RGB with a resolution of 2464$\times$2056 pixels.

Furthermore, camera calibration was carried out by capturing 20 calibration images for each group. Each of the images contains a checkerboard with known dimensions (each square is $20.0$ x $20.0$ mm) in various poses, including placing the checkerboard flatly on the conveyor belt.

\subsection{Image collection}
\label{sec:img_collection_and_annotation}
Prior to capturing the images for the dataset, the length of every fish was measured by an experienced marine biologist, following common practice where the number is rounded to the nearest 5 mm.
Next, the fish were partitioned into 25 groups, with 14 to 24 fish in each group.
As opposed to \cite{GarciadUrso2022}, where the fish appear sorted according to species on the tray, we selected the number of fish and distribution of species in each group pseudo-randomly to mimic real-world scenarios where the fish are processed together when hauled onto the fishing vessel.

The groupings and fish IDs allow us to capture multiple images of the same fish in different orientations and positions, while systematically ensuring that every fish is represented by the same number of images.
Additionally, the groups make it convenient to create training and test splits without data cross-contamination.

\begin{figure}[t]
    \centering
    \includegraphics[width=\linewidth]{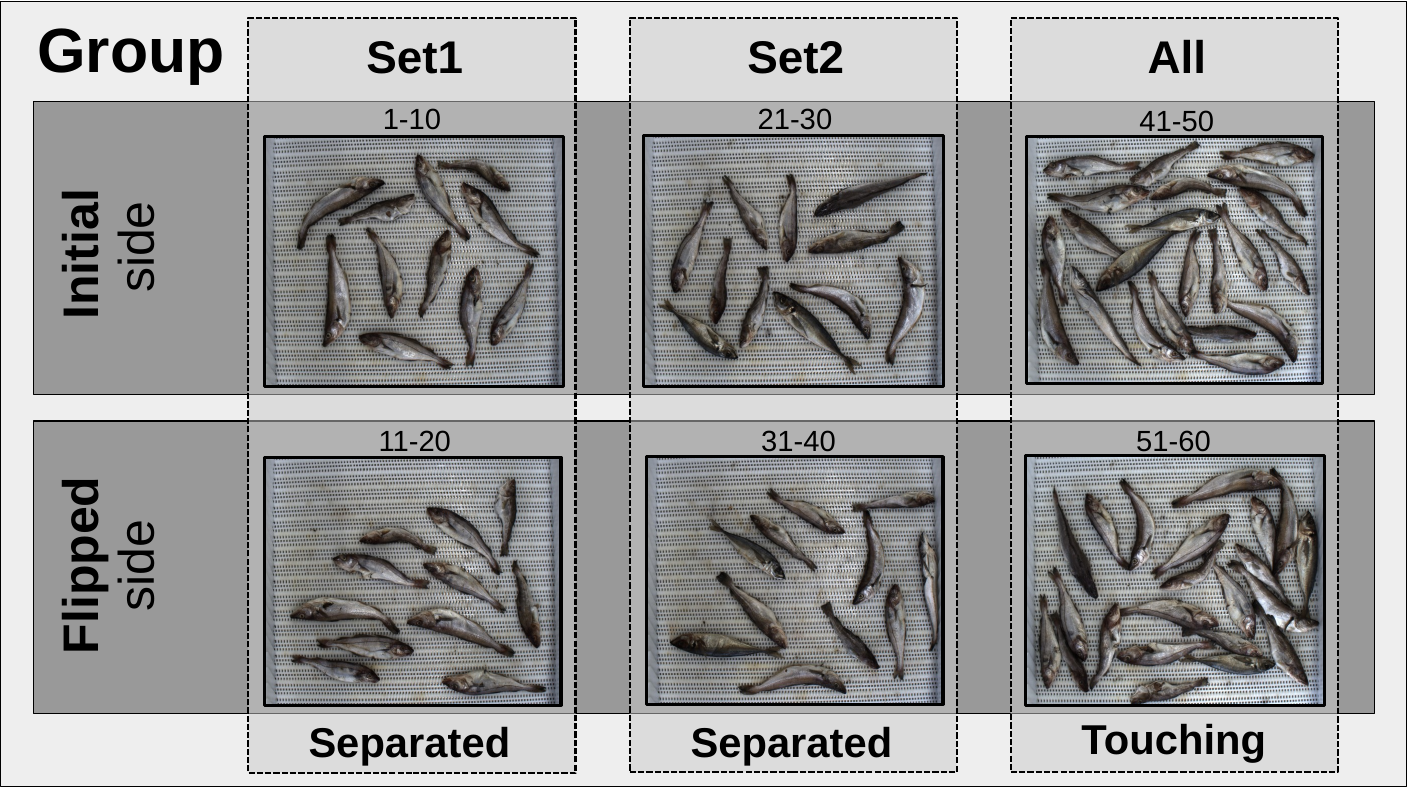}
    \caption{The AutoFish dataset contains 25 groups of fish. Each group consists of three subsets of images, namely, \textit{Set1} and \textit{Set2}, which contain one half of the fish each, and \textit{All}, which contains all of the group's fish.}
    \label{fig:collection}
\end{figure}

Every group is partitioned into three subsets: \textit{Set1}, \textit{Set2}, and \textit{All}, each of which comprises 20 images, as illustrated in~\cref{fig:collection}. 
\textit{Set1} and \textit{Set2} contain half of the fish each, and none of the fish can overlap or touch each other. 
On the other hand, \textit{All} contains all the fish in the group, purposely placed in positions where they touch and occlude each other. 
In the total 60 images of a group, every fish will appear exactly 40 times, with 20 times from each side.
This gives us varying levels of difficulties with respect to detection, segmentation, length estimation, and other downstream tasks.

\subsection{Annotation procedure}
During recording, every fish is meticulously point-annotated with its ID in every image, before shuffling them on the table and repeating the procedure.
This allows us to keep track of all fish throughout the session. 
We provide accurate instance segmentation masks for individual fish in every image of the dataset.
We have used the open-source and Python-based annotation software Labelme\footnote{\url{https://github.com/labelmeai/labelme}} for annotating the images.
First, we leveraged a Segment Anything Model (SAM)~\cite{kirillov2023segany} to obtain initial segmentation masks based on the manual point-annotations acquired during the image-acquisition procedure.
The SAM-based masks were then inspected and manually corrected to ensure the best possible fit.
Lastly, in cases where occlusion caused single fish to appear in multiple masks, the masks were associated with the same ID. 
All annotations were compiled in a single file following the MS COCO \cite{lin2015microsoftcococommonobjects} format.

\section{Methods}\label{sec:methods}
Automated catch documentation processes rely on species identification and length estimation. 
Since fish are non-rigid and can adopt irregular shapes, segmentation masks offer a more precise and visually intuitive method for delineating object boundaries compared to alternatives like bounding boxes or keypoints. 
This makes masks especially valuable for downstream tasks requiring human inspection and verification. 
In this work, we generate instance segmentation masks and demonstrate their use in estimating fish lengths through a two-stage process.

\subsection{Instance segmentation}
In our experiments, we use the instance segmentation model Mask2Former~\cite{cheng2021mask2former} to provide baseline results.
Two variations of the architecture are deployed: a configuration equipped with a traditional convolutional ResNet-50 \cite{7780459} backbone and a larger alternative, using a transformer Swin-base \cite{9710580} (Swin-B) backbone. 
All instances of the models are pre-trained on MS COCO \cite{lin2015microsoftcococommonobjects} and fine-tuned for 1000 steps without a validation set or early stopping.
The batch size is set to 8 images.
Following the default configuration of Mask2Former, the optimizer used is \textit{ADAMW} \cite{loshchilov2019decoupledweightdecayregularization}.
The learning rate is controlled through a multistep scheduler from 0.1 to 0.0001.

During training, we apply common image augmentations and the hyperparameters are outlined in \cref{tab:hyperparams}.
Random horizontal and vertical flips, with probabilities $p_h$ and $p_v$, respectively, are applied to make it less likely for the model to learn certain positions and orientations of the fish.
As the scene was lit in part by uncontrolled natural light, there is a slight difference between some of the images depending on the time of day, the weather, and more.
To minimize the impact of light variation, we introduce contrast ($c$), brightness ($b$), and saturation ($s$) augmentations during training.

\begin{table}[b]
\centering
\resizebox{\columnwidth}{!}{%
\begin{tabular}{p{55pt}ccccc}
\toprule
\textbf{Task}         & $\textbf{p}_\textbf{h}$ & $\textbf{p}_\textbf{v}$ & $\textbf{c}$ & $\textbf{b}$ & $\textbf{s}$  \\ \midrule
\textbf{Segmentation}          & 0.5            & 0.5            & [0.75;1.25]    & [0.75;1.25]    & [0.75;1.25]     \\
\textbf{Length est.}           & 0              & 0              & [0.50;1.50]    & [0.80;1.20]    & [0.60;1.40]     \\
\bottomrule
\end{tabular}}
\caption{Augmentation hyperparameters for the proposed instance segmentation and length estimation models.}
\label{tab:hyperparams}
\end{table}

\subsection{Length estimation}
For providing baseline fish length estimations, we implement and evaluate two approaches.
The first, denoted \textit{SKL}, relies on classic image processing techniques in the form of applying skeletonization on the instance segmentation masks.
The second, denoted \textit{REG}, is a learning-based approach utilizing a small convolutional neural network for length regression.

\subsubsection{Mask skeletonization (SKL)}
\begin{figure}[t]
\centering
    \begin{subfigure}{0.95\columnwidth}
        \includegraphics[width=0.99\linewidth]{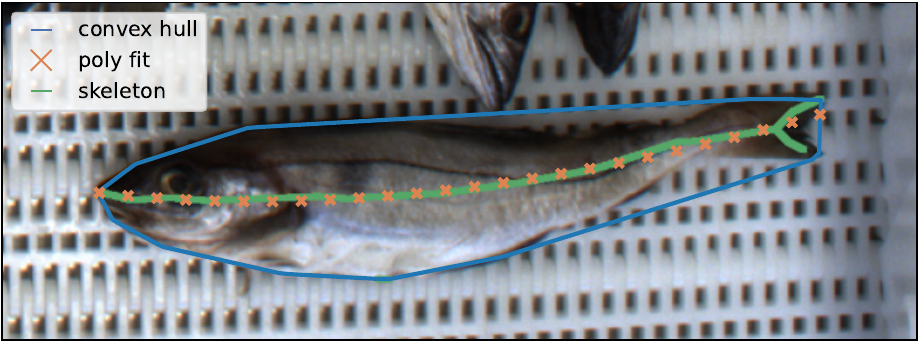}
    \end{subfigure}
    \begin{subfigure}{0.95\columnwidth}
        \includegraphics[width=0.99\linewidth]{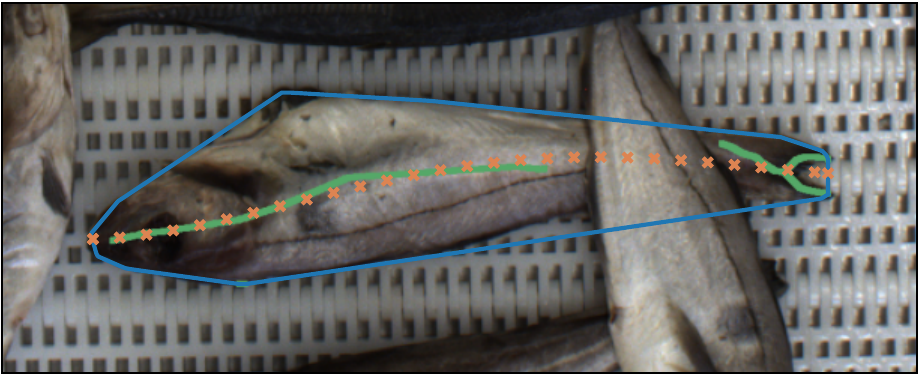}
    \end{subfigure}
\caption{The central line is identified by fitting a polynomial (orange) to the skeleton of the mask (green). Secondly, the polynomial is evaluated based on the convex hull of the mask (blue) to handle forked caudal fins and occlusions.}
\label{fig:skeletonizaiton_examples}
\end{figure}

The first step of the skeletonization-based length estimation is to determine the pixel-wise length of each fish along its central line, as illustrated in \cref{fig:skeletonizaiton_examples}.
We do this by processing the segmentation masks using the skeletonization method proposed by Zhang et al.~\cite{zhang1984fast}.
To approximate a smooth central line of the fish, from which the length can be estimated, we fit a 4th-degree polynomial to the skeleton of the mask.
To account for forked caudal fins and occlusions, where the mask may be split into multiple segments, we compute the convex hull of the mask and evaluate the polynomial along the boundaries of the convex hull.

The next step is to obtain a mapping from image plane (pixels) to the surface of the conveyor belt (centimeters).
This is done by estimating a homography for every group of fish based on 20 calibration images, which are also used for correcting lens distortion in the images.
Finally, the length in centimeters of each fish can be estimated by taking the image points from the polynomial fit and mapping them onto the plane of the conveyor belt and accumulating the distance between them. 

\subsubsection{CNN-based length regression (REG)}
\begin{figure}[t]
\centering
        \includegraphics[width=0.99\linewidth]{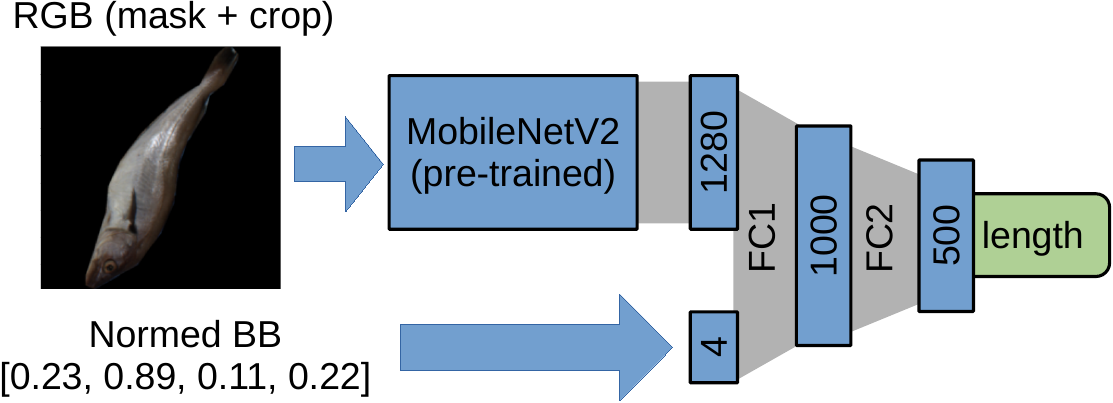}
\caption{Overview of the CNN-based regression model (REG).}
\label{fig:cnn_length_est_pipeline}
\end{figure}

This approach is inspired by the work of Ovalle et al.~\cite{Ovalle2022}, which showed that a small MobileNetV1 with a regression head was sufficient for estimating the length of fish on a conveyor belt. 
We use an ImageNet~\cite{Deng2009} pre-trained MobileNetV2~\cite{Sandler2018} model with a custom regression head consisting of two fully connected layers, as shown in \cref{fig:cnn_length_est_pipeline}.

The masks from the instance segmentation network goes through a pre-processing step before being fed into the network. 
The input image is cropped to fit a black squared bounding box around the RGB mask of the fish, before it is fed into the MobileNetV2 model.
Information regarding the spatial position of the object is critical for the pixel to centimeter mapping.
Therefore, the normalized bounding box coordinates are provided to the fully connected layers in addition to the embedded image features.

The entire model, including the MobileNetV2 backbone, is trained using a batch size of 32 for a total of 200 epochs.
The model is trained using the \textit{L1} loss along with the \textit{ADAM} optimizer using a fixed learning rate of $0.001$.
During training, randomized image augmentations were used, as reported in \cref{tab:hyperparams}.
No geometric transformations were applied as they may affect the pixel to centimeter mapping.

\section{Results}\label{sec:results}
The \textit{Autofish} dataset is divided into groups to support various split configurations while preventing data cross-contamination between the splits.
However, for the evaluation, the following five groups are reserved as the test split: [10, 14, 20, 21, 22] and are excluded from all training. 
These groups were selected to reflect the overall class distribution of the dataset.

\subsection{Instance segmentation}
The performance of our instance segmentation models are evaluated based on the mean average precision, which is calculated as $mAP = AP@[IoU = .5:.95]$ by thresholding the intersection over union (IoU) between the predictions and the ground truth annotations in steps of 0.05.

We evaluate the models per class on the \textit{separated}, \textit{touching}, and \textit{combined} image sets, both training and testing each configuration 10 times with different initializations. 
The resulting mAPs and error margins are presented in \cref{tab:instance_segmentation_results}.
The performance of both \textit{Mask2Former} models are consistent, with a small error margin of $<1\%$. 
Swin-B outperforms the ResNet50 backbone across all species and image categories.

\begin{figure}[t]
\centering
\includegraphics[width=\columnwidth]{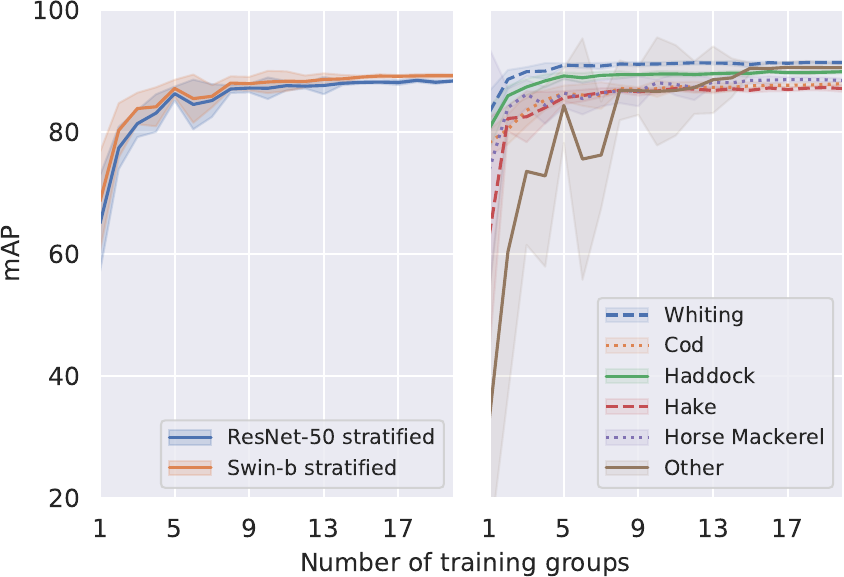}
\caption{Model mAP and standard deviation as the number of training groups is gradually increased. On the left, a comparison between the two backbones. On the right, the per-class results for the Swin-b backbone.}
\label{fig:varying_training_size}
\end{figure}

To analyze the impact of the training split size, the models were trained on 20 progressively larger random subsets of the training data.
For each subset size (except 20) 10 random variations of groups were used for training the models.
The results are presented in \cref{fig:varying_training_size}, which shows that both models stabilize at 9 groups ($\sim180$ different fish and $\sim540$ images), reaching almost maximum performance with very little variation due to the specific groups used for training.
This indicates the possibility of relatively easy and cheap adaptations to the expected catch of individual vessels. 

\begin{table*}[b]
\centering
\begin{tabular}{lcccccccc}
 & \multicolumn{2}{c}{Separated} & & \multicolumn{2}{c}{Touching} & & \multicolumn{2}{c}{Combined} \\ \cline{2-3}\cline{5-6}\cline{8-9} 
 \multicolumn{1}{c}{} & ResNet50       & Swin-B       && ResNet50      & Swin-B     && ResNet50    & Swin-B    \\ \hline
All                                        & $92.47\pm0.12$               & $\textbf{92.98}\pm\textbf{0.21}$             &&  $84.09\pm0.24$             & $\textbf{85.32}\pm\textbf{0.32}$           && $88.31\pm0.13$            & $\textbf{89.15}\pm\textbf{0.26}$          \\ \hline
Whiting                                    & $93.13\pm0.27$               & $94.20\pm0.28$             &&  $87.99\pm0.37$             & $88.69\pm0.34$           && $90.47\pm0.26$            & $91.32\pm0.27$          \\
Cod                               & $91.09\pm0.13$               & $91.44\pm0.24$             &&  $83.27\pm0.62$             & $83.97\pm0.47$           && $87.21\pm0.28$            & $87.86\pm0.29$          \\
Haddock                                    & $91.73\pm0.28$               & $92.63\pm0.31$             &&  $85.78\pm0.24$             & $86.94\pm0.37$           && $88.75\pm0.20$            & $89.82\pm0.26$          \\
Hake                                       & $90.55\pm0.35$               & $90.85\pm0.59$             &&  $82.18\pm0.46$             & $83.16\pm0.38$           && $86.49\pm0.36$            & $87.06\pm0.54$          \\
Horse mackerel                             & $91.85\pm0.28$               & $92.29\pm0.21$             &&  $82.74\pm0.82$             & $84.52\pm0.65$           && $87.29\pm0.48$            & $88.38\pm0.28$          \\
Other                                      & $96.49\pm0.34$               & $96.50\pm0.43$             &&  $82.57\pm0.93$             & $84.65\pm0.97$           && $89.66\pm0.35$            & $90.48\pm0.54$          \\ \hline
\end{tabular}
\caption{Instance segmentation results for the ResNet50 and Swin-B backbones when trained on all 20 training groups. The mAP and error margin are based on 10 random model initializations.}
\label{tab:instance_segmentation_results}
\end{table*}

\begin{figure}[t]
\centering
    \begin{subfigure}{0.49\columnwidth}
        \includegraphics[width=\linewidth]{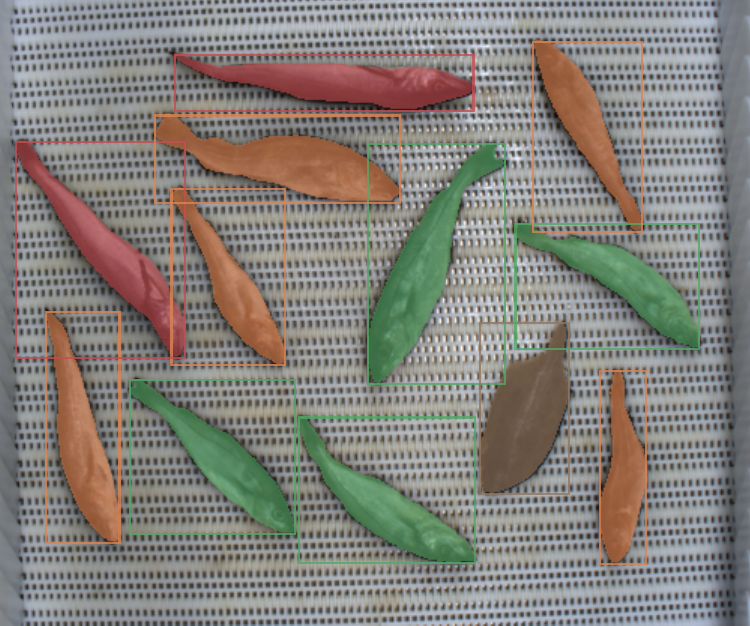}
        \caption{mAP = 0.96}
    \end{subfigure}
    \begin{subfigure}{0.49\columnwidth}
        \includegraphics[width=\linewidth]{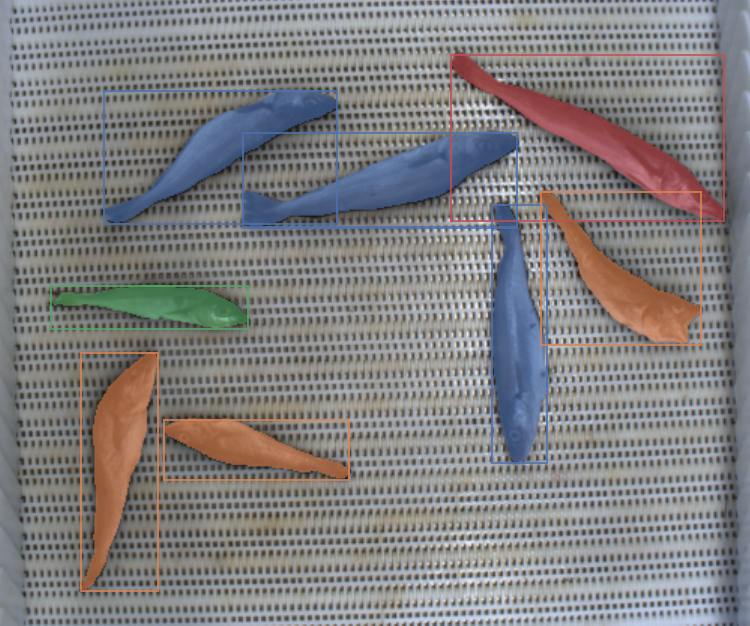}
         \caption{mAP = 0.97}
    \end{subfigure}
    
    \begin{subfigure}{0.49\columnwidth}
        \includegraphics[width=\linewidth]{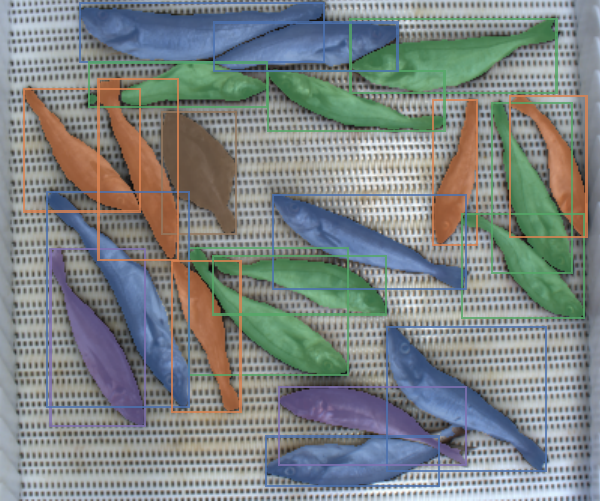}
        \caption{mAP = 0.94}
    \end{subfigure}
    \begin{subfigure}{0.49\columnwidth}
        \includegraphics[width=\linewidth]{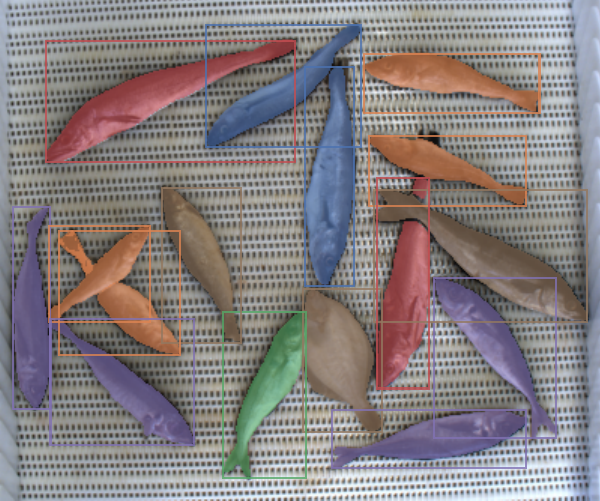}
         \caption{mAP = 0.91}
    \end{subfigure}

    \begin{subfigure}{0.49\columnwidth}
        \includegraphics[width=\linewidth]{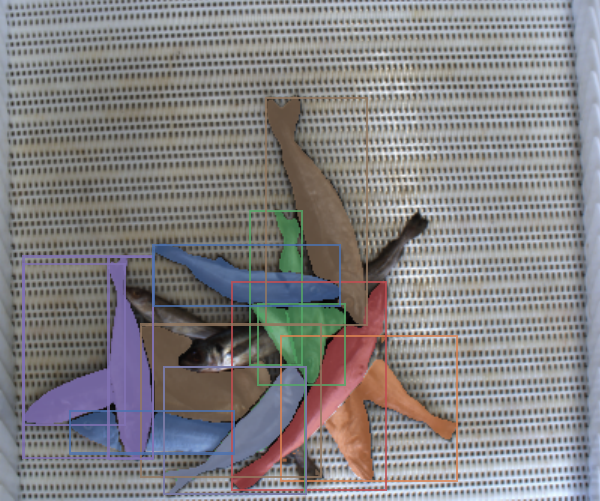}
         \caption{mAP = 0.37}
    \end{subfigure}
    \begin{subfigure}{0.49\columnwidth}
        \includegraphics[width=\linewidth]{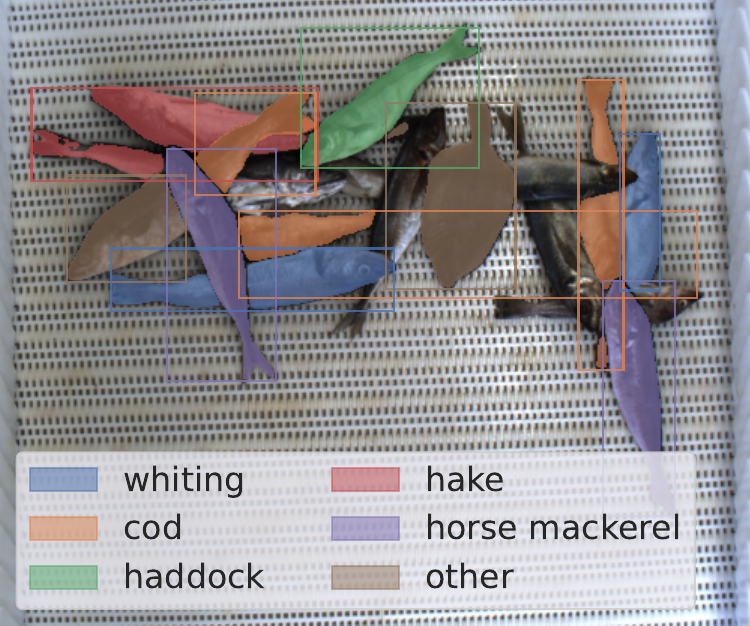}
         \caption{mAP = 0.52}
    \end{subfigure}
\caption{Model output examples. Instances of good performance are displayed on the top four images, while low scores are on the bottom. All labels have been predicted correctly.}
\label{fig:example_predictions}
\end{figure}

Characteristic examples of the model's detection capabilities are presented in \cref{fig:example_predictions}.
Images in the top have a high mAP, close to the general performance of the models, while the images on the bottom feature some of the worst cases.
The confidence threshold chosen for the output displayed is $0.9$, as our analysis showed that 96\% of matched predictions fall above it, with an average mask IoU of 0.94.

\subsection{Length estimation}
The length estimation methods are evaluated based on the mean absolute error (MAE) in centimeters and the mean absolute percentage error (MAPE). 
Additionally, the methods are evaluated with two types of input: the Swin-B predictions (denoted \emph{pd}) and the ground truth masks (denoted \emph{gt}). 
For the \emph{pd} masks, only predictions exceeding a confidence threshold of $90\%$ were used.
The training groups were split into a train and a validation split, with 15 and 5 groups, respectively. 
The validation split was randomly picked to contain the following groups: $[1, 6, 11, 17, 25]$.

\begin{table}
\centering
\begin{tabular}{lccc}
\toprule & Separated             & Touching                 & Combined \\ \midrule
SKL$^{gt}$  & \textbf{0.59 (1.79\%)} & 1.43 (4.51\%)            & 1.01 (3.15\%) \\
REG$^{gt}$   & 0.67 (2.10\%)          & \textbf{0.96 (3.08\%)}   & \textbf{0.82 (2.59\%)} \\
\rowcolor[HTML]{EFEFEF} 
SKL$^{pd}$   & \textbf{0.62 (1.87\%)} & 2.43 (7.42\%)            & 1.51 (4.59\%) \\
\rowcolor[HTML]{EFEFEF} 
REG$^{pd}$   & \textbf{0.62} (1.92\%) & \textbf{1.38 (4.32\%)}   & \textbf{0.99 (3.10\%)} \\\hline

\end{tabular}
\caption{Results for the length estimation reported as MAE in centimeters and MAPE. The skeletonization-based (\emph{SKL}) and regression-based (\emph{REG}) methods are evaluated using both groundtruth masks (\emph{gt}) and actual predicted masks (\emph{pd}).}
\label{tab:length_mae_results}
\end{table}

The performance of the two length estimation approaches are summarized in Table \ref{tab:length_mae_results}, where we see that the skeletonization method (SKL) achieves the best performance on the images from the \textit{separated} sets.
However, in the more challenging sets where the fish are allowed to touch and occlude each other, the CNN regression-based model (REG) is significantly better.

We present the error distributions of the two approaches in \cref{fig:length_histogram_results} as histograms, along with the corresponding mean and standard deviation. 
The notable spikes at the boundaries of the histograms are due to the errors being clipped to a max of $\pm 5.0$ centimeters and hence being accumulated in the outer bins in the histogram.
This is done on purpose to avoid a few outliers skewing the plots and also to clearly show the proportion of errors outside this range for the different scenarios.

\section{Discussion}\label{sec:discussion}
The presented baseline experiments show that with state-of-the-art deep neural network architectures, it is possible to automate fish identification on conveyor belts to a large degree.
When not presented with heavy occlusion, the models are very consistent at differentiating fish species, even when they look indiscernible to a nonspecialist. 
In cases with high levels of occlusion, e.g., where fish overlap to the extent where individual fish are divided into multiple segments, the models are able to predict these segments and correctly classify them as belonging to the same individual.
It is only in the very extreme cases that the models miss fish entirely.
Additionally, the predicted masks are generally of significant quality, with an expected IoU over 0.85. 
This is critical for the performance of further downstream tasks, like skeletonization, that directly depend on the quality of the mask. 

During the evaluation of the length estimation it was found that the skeletonization and CNN-based approaches had a similar performance of $0.62$ cm MAE in scenarios with no occlusion (\emph{separated}), as shown in \cref{tab:length_mae_results}.
This suggests that a length estimation approach based on classic image processing, such as skeletonization, is a viable option if there is a low risk of occlusion.
This could be valuable for systems mounted on smaller fishing vessels with limited power supply.
In the sets where the fish are touching or occluding each other, \emph{touching} and \emph{combined}, the CNN-based approach clearly outperforms the skeletonization-based approach. 
This is not unexpected, as the skeletonization-based approach is not able to account for occlusions that causes the head or caudal fin of the fish to be missing from the mask. 
Therefore, it will have a tendency to underestimate the length of the fish, which is clearly seen in the histograms for the error distribution in \cref{fig:length_histogram_results}.
The CNN-based approach, on the other hand, can learn to infer the length based on other features of the mask, in cases where the head or caudal fin are not visible.

\newcommand{\tabimage}[1]{\adjustbox{valign=c,vspace=1pt}{\includegraphics[width=.46\linewidth]{#1}}}
\begin{figure}[t]
\centering
\addtolength{\tabcolsep}{-0.5em}
\begin{tabular}{lcc}
       & Groundtruth & Predicted \\
\rotatebox[origin=c]{90}{Separated}  
& \tabimage{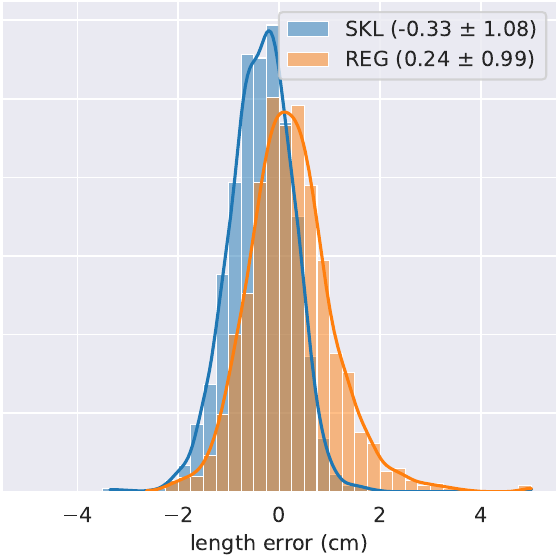} 
& \tabimage{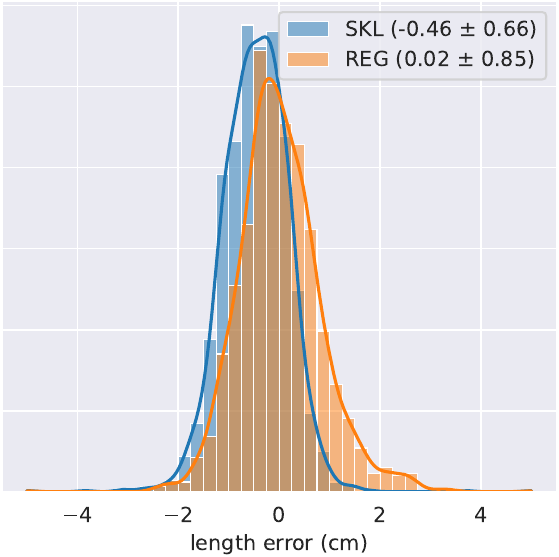} \\
\rotatebox[origin=c]{90}{Touching}  
& \tabimage{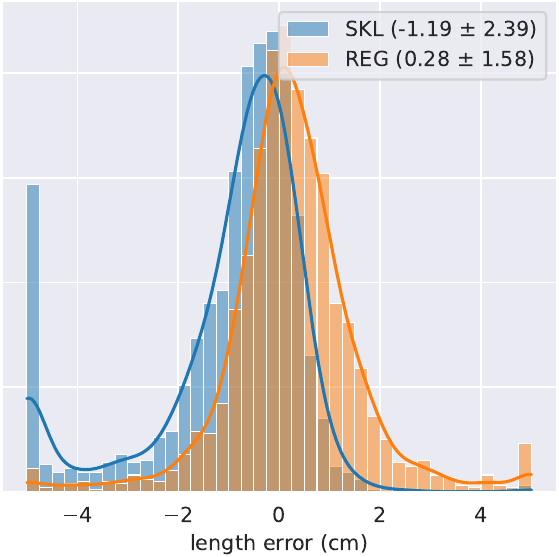} 
& \tabimage{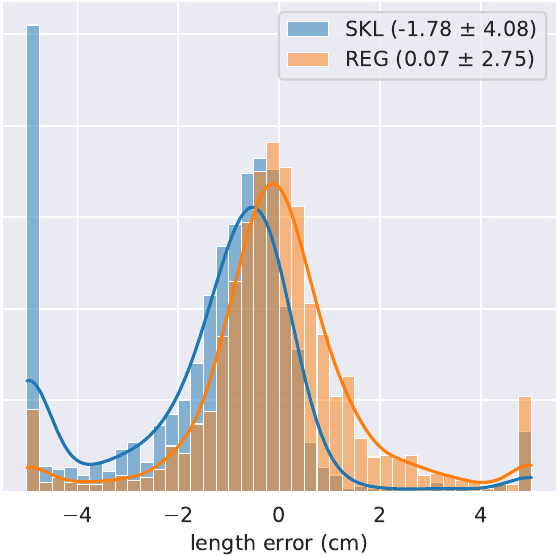} \\ 
\rotatebox[origin=c]{90}{Combined}  
& \tabimage{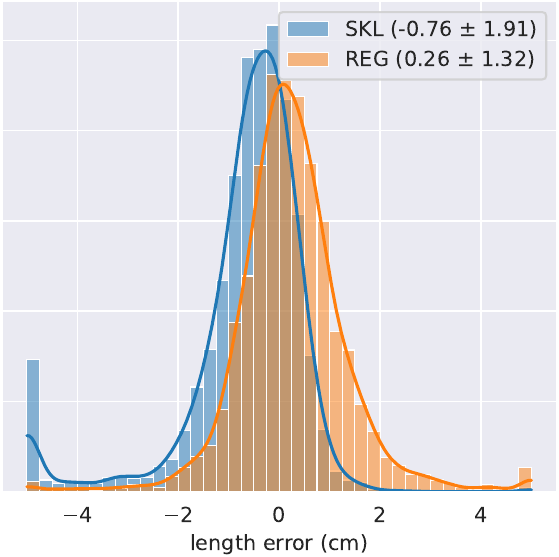} 
& \tabimage{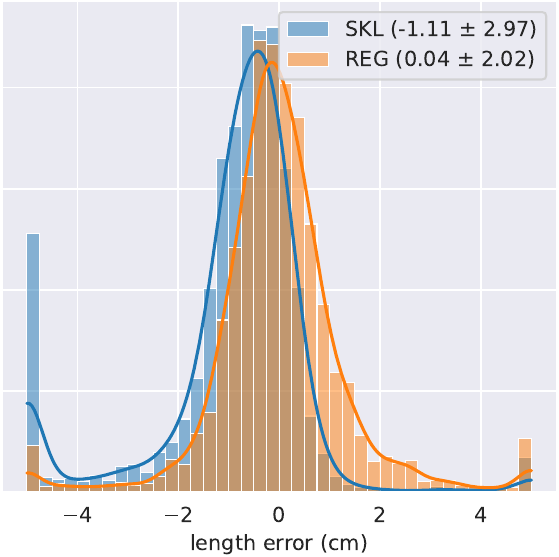} \\ 
\end{tabular}
\caption{The distribution of length estimation error in centimeters for both the skeletonization-based (\emph{SKL}) and CNN-based (\emph{REG}) approach. All errors have been limited to max $\pm 5.0$ centimeters to avoid outliers skewing the plots. Both the mean error and standard deviation are reported for each approach in the legend of the respective plot.}
\label{fig:length_histogram_results}
\end{figure}

A practical distinction between the two length estimation methods is that the skeletonization requires only a few calibration images to adapt to a new setup. 
In contrast, the CNN necessitates new training data and additional training to be deployed in another environment.
To summarize, the skeletonization is easier to adapt to new setups, but struggles with occluded objects.
The CNN, on the other hand, handles occlusions effectively but demands new training to adapt to new setups, making it more resource-intensive.

\subsection{Actively using IDs in future work}
\begin{figure}[t]
\centering
    \includegraphics[width=0.95\linewidth]{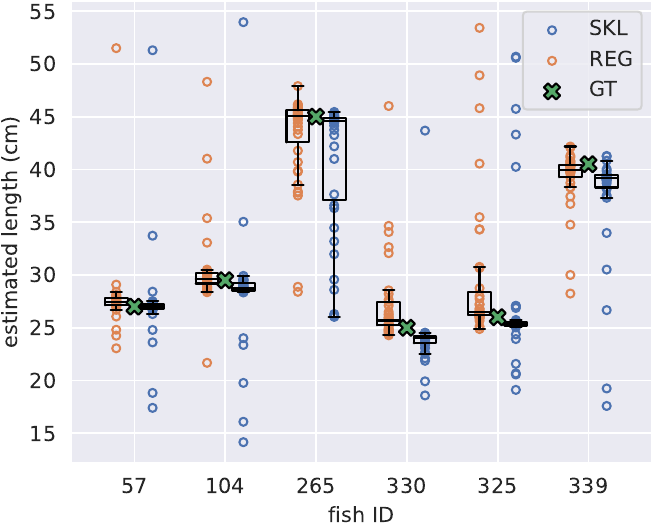}
\caption{Boxplots for the six fish with the highest variation in their estimated lengths. The actual estimated lengths are plotted for each fish for both the skeletonization-based (\emph{SKL}) and CNN-based (\emph{REG}) approaches. The manually measured ground truth lengths are plotted as well (\emph{gt}).}
\label{fig:length_est_boxplot_highest_std}
\end{figure}

\begin{figure}[t]
\centering
    \includegraphics[width=0.95\linewidth]{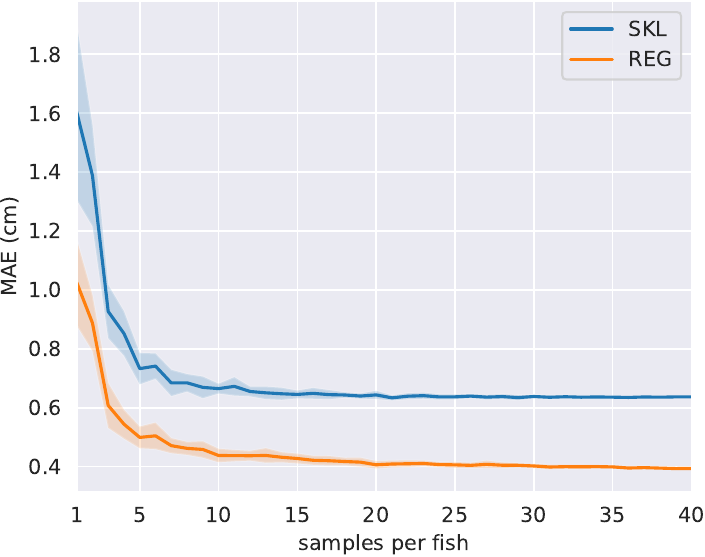}
\caption{MAE as a function of the number of samples using the median of the length estimation for each fish. The shaded areas denotes the standard deviation of the MAE as it changes depending on which samples are available for the averaging of the lengths. The plot is based on the predicted masks and the \emph{combined} set.}
\label{fig:mae_vs_samples}
\end{figure}

A particular aspect of the \emph{AutoFish} dataset is the inclusion of unique IDs for every fish.
This allows for conducting fine-grained analyses of the results, such as the length estimation, on a fish-by-fish level.
Individual fish associated with large variations in their estimated lengths were identified and are shown with boxplots in \cref{fig:length_est_boxplot_highest_std}.
The samples highlight that there can be a vast difference in the length estimations, even for the CNN-based approach (REG), that is able to handle occlusion to some degree.
However, when looking at the median estimated length, both methods  are close to the actual ground truth measurement.
This suggests that length estimation could benefit from the use of multiple samples for the same fish to increase accuracy.
To see how many samples are needed to achieve satisfactory length estimations, the MAE metric was re-calculated using the median estimated length across each individual fish ID with a varying number of samples.
The results are plotted in \cref{fig:mae_vs_samples} and simulate a scenario where it is possible to maintain IDs for the individual fish, either through tracking or using re-identification.

Both length estimation approaches appear to benefit from having more samples per fish, but the performance boost is more significant for the CNN-based approach.
Identifying 40 samples per fish could potentially boost the MAE from $0.99$ cm down to $\approx 0.4$ cm for the CNN-based approach.
In cases when such a high number of samples per fish is not feasible, having even five samples could still reduce the MAE to $\approx 0.5$ cm.
This result is at the limit in terms of the precision of the ground truth lengths, as each fish was manually measured to the nearest 5 mm.

Investigating whether it is feasible to maintain fish IDs through re-identification is a promising option for future work with the \emph{AutoFish} dataset.
Especially considering that the data are already readily available in the form of 40 images for each of the 454 fish in the dataset.
The re-identification task is not only relevant for improving length estimation accuracy, but could also allow for the documentation of fish at an individual level as they are caught or within processing facilities. 

\section{Ethical statement} \label{sec:ethical}
Fish used in these experiments were caught and landed by fishermen following relevant legislation and normal fishing procedures.
The Danish Ministry of Food, Agriculture and Fisheries of Denmark was contacted before fish collection to ensure compliance with legislation.
The fish were dead at landing and only dead fish were included in this experiment.
There is no conflict with the European Union (EU) directive on animal experimentation (article 3, 20.10.2010, Official Journal of the European Union L276/39) and Danish law (BEK nr 12, 07/01/2016).
The laboratory facilities used at Aalborg University are approved according to relevant legislation. 

\section{Acknowledgment} \label{sec:acknowledgments}
The project is financed by the European Union, the European Maritime and Fisheries Fund (EMFF) and the Danish Agricultural and Fisheries Agency (AUTOFISK-33113-I-20-175).
We would like to thank Helle Blendstrup, Poul Lund, and Alex Jørgensen for their invaluable help.

%%%%%%%%% REFERENCES
{\small
\bibliographystyle{ieee_fullname}
\bibliography{egbib}
}

\end{document}